\documentclass[letterpaper]{article} 
\usepackage{aaai2026}  
\usepackage{times}  
\usepackage{helvet}  
\usepackage{courier}  
\usepackage[hyphens]{url}  
\usepackage{graphicx} 
\usepackage{natbib}  
\usepackage{caption} 
\usepackage{bm}
\usepackage{amsmath}
\usepackage{subcaption}
\usepackage{amssymb}
\usepackage{algorithm}
\usepackage{algorithmic}

\usepackage{newfloat}
\usepackage{listings}
\DeclareCaptionStyle{ruled}{labelfont=normalfont,labelsep=colon,strut=off} 
\floatstyle{ruled}
\newfloat{listing}{tb}{lst}{}
\floatname{listing}{Listing}
\title{Improving the Robustness of Control of Chaotic Convective Flows with Domain-Informed Reinforcement Learning}
\author {
    Michiel Straat\textsuperscript{\rm 1},
    Thorben Markmann\textsuperscript{\rm 1},
    Sebastian Peitz\textsuperscript{\rm 2},
    Barbara Hammer\textsuperscript{\rm 1}
}
\affiliations {
    \textsuperscript{\rm 1}Center for Cognitive Interaction Technology (CITEC), Bielefeld University, Inspiration 1, 33619 Bielefeld, Germany\\
    \textsuperscript{\rm 2}Department of Computer Science \& Lamarr Institute for Machine Learning and Artificial Intelligence,\\TU Dortmund University, Joseph-von-Fraunhofer-Straße 25, 44227 Dortmund, Germany\\
    \{mstraat, tmarkmann, bhammer\}@techfak.uni-bielefeld.de, sebastian.peitz@tu-dortmund.de
}

\usepackage{bibentry}

\newcommand{\Ra}{\text{Ra}}
\newcommand{\Nu}{\text{Nu}}

\begin{document}

\maketitle

\begin{abstract}
Chaotic convective flows arise in many real-world systems, such as microfluidic devices and chemical reactors. Stabilizing these flows is highly desirable but remains challenging, particularly in chaotic regimes where conventional control methods often fail. Reinforcement Learning (RL) has shown promise for control in laminar flow settings, but its ability to generalize and remain robust under chaotic and turbulent dynamics is not well explored, despite being critical for real-world deployment.
In this work, we improve the practical feasibility of RL-based control of such flows focusing on Rayleigh-B\'enard Convection (RBC), a canonical model for convective heat transport.
To enhance generalization and sample efficiency, we introduce domain-informed RL agents that are trained using Proximal Policy Optimization across diverse initial conditions and flow regimes. We incorporate domain knowledge in the reward function via a term that encourages B\'enard cell merging, as an example of a desirable macroscopic property.
In laminar flow regimes, the domain-informed RL agents reduce convective heat transport by up to 33\%, and in chaotic flow regimes, they still achieve a 10\% reduction, which is significantly better than the conventional controllers used in practice.
We compare the domain-informed to uninformed agents: Our results show that the domain-informed reward design results in steady flows, faster convergence during training, and generalization across flow regimes without retraining.
Our work demonstrates that elegant domain-informed priors can greatly enhance the robustness of RL-based control of chaotic flows, bringing real-world deployment closer.
\end{abstract}

\begin{links}
    \link{Code}{https://github.com/HammerLabML/RBC-Control-SARL}
    \link{Videos}{}In the file examples.md in the code repository
\end{links}

\section{Introduction}
AI methods hold great promise for advancing engineering applications in fluid dynamics \cite{wang_2024}, including those in industry, aviation, energy systems, and climate science. Traditionally, these domains rely on numerical simulations of physical laws, commonly the Navier-Stokes equations, for tasks ranging from airfoil design to weather prediction. With the improvements in sensors and computational capacity, AI now enables data-driven approaches to modeling and control, opening new opportunities to tackle complex flow control problems \cite{vinuesa_transformative_2023}.

\textit{Natural convection} is the process of heat transport by fluid motion due to temperature differences. It is common in nature and industry, and its controllability plays a crucial role in process reliability and quality.
For example, in crystal growth processes or microfluidic gradient generation, uncontrolled convection can introduce instabilities that degrade material quality or disrupt precision measurements \cite{tang_1993, gu2018}.
In this work, we study robust RL-based control of Rayleigh-B\'enard Convection (RBC), a canonical model for convective heat transport that captures the transition from laminar to chaotic flow as temperature differences increase.

Conventional control methods often struggle to stabilize such chaotic systems. In contrast, Reinforcement Learning (RL) has shown growing promise in flow control tasks such as turbulence suppression, mixing optimization, and drag reduction \cite{garnier_2021}.
However, the robustness and generalization of RL-based control of chaotic flows is largely underexplored. This currently limits real-world deployments, where small variations in initial conditions or system parameters can lead to widely different behaviors.
In the context of convective dynamics, it is especially important for control agents to generalize across both initial conditions and different chaotic regimes, ideally without requiring retraining.

In AI for scientific applications, incorporating prior knowledge, such as conservation laws or physical symmetries, has helped improve sample efficiency, generalization~from scarce and noisy data, and physical plausibility of learned models \cite{banerjee_2025}.
Inspired by this, we explore a domain-informed reward shaping approach, where we embed physically meaningful macroscopic features into the reward function. We show that this addition holds promise to guide the agent toward flow stabilization strategies that generalize across flow regimes and initial conditions, and require fewer rollouts during training.

Initial works have explored the control of convection by RL: \citet{vignon_2023,vasanth2024} proposed a scalable RL approach for a laminar flow regime in 2D and 3D, but the works did not address generalization across initial conditions or system parameters. \citet{beintema_2020} studied control of chaotic convective flows in a narrow domain and identified promising emergent strategies, but the work did not explore robustness across flow regimes and did not leverage domain knowledge.

Motivated by the open challenge of robust control for turbulent convection, we make three key contributions:
\begin{enumerate}
\item We introduce a domain-informed reward that reduces training time and promotes fast flow stabilization.
\item We demonstrate that the domain-informed reward improves generalization across initial conditions and chaotic flow regimes.
\item We achieve robust control in considerably chaotic convective flow settings where previous methods either fail or have not been tested.
\end{enumerate}
These contributions move RL-based flow control for chaotic flows a step closer to practical deployment.
Alongside each section and experiment, we provide videos in the code repository (link below abstract). We also provide all experiment details there and in the Supplementary Material (henceforth denoted as SM).

\section{Methodology} \label{sec:methods}
In this section, we present the necessary details of the dynamical system, its parameters, and measurement quantities. We then introduce the control task and our training setup. 

\subsection{Rayleigh-B\'enard Convection} \label{sec:methods:rbc}
Rayleigh-B\'enard Convection (RBC) is a widely studied model in fluid dynamics for heat transport between a heated lower plate and a cooled upper plate, see e.g. \cite{pandey2018} for the partial differential equation.

We consider RBC on a 2D domain with horizontal coordinate $x \in [0,2\pi]$ and vertical coordinate $y \in [-1,1]$, i.e. our domain has width $W=2\pi$ and height $H=2$.
The system's state in 2D consists of the velocity field $\bm{u}(x,y)=(u_x, u_y)$ and the scalar temperature field $T(x,y)$. We use periodic boundaries at the left and right of the domain. For the bottom and top boundaries, we use zero-velocity (i.e. no-slip) conditions and for the temperature we use $T_b$ at the bottom and $T_t$ at the top, where $T_b > T_t$ drives the convective flow.
We used a spectral solver from the Shenfun package \cite{mortensen_joss} on a grid of collocation points of size $96 \times 64$. \footnote{All other simulation parameters can be found in the Supplementary Material.}
The main system parameter is the \textit{Rayleigh number} (Ra). It is a ratio of the timescale of diffusive to convective thermal transport, i.e. higher Ra means more convection.
Interesting dynamics occur in the system in dependence of Ra: for low Ra, the fluid does not move and the temperature field converges to a stable equilibrium through heat conduction only, which is a linear temperature gradient in the vertical direction:
\begin{equation}
T_{cond}(y)=T_b-y/H(T_b-T_t)\, .
    \end{equation}
As Ra increases past a critical value, the system becomes unstable and heat starts to be transported by convective flow, i.e. hot fluid rises, and cold fluid sinks which typically takes place in the form of so-called \textit{B\'enard cells} (see Fig.~\ref{fig:example_ra10000}).
As Ra increases further, the fluid flow becomes increasingly chaotic and eventually turbulent \cite{hsia_route_2022}.
In this regime, the dynamics depend sensitively on initial conditions and the value of Ra.
We selected a range of $\Ra$ that includes unsteady flows and moderately turbulent effects, which make for an interesting benchmark for robust learning-based control.

\subsubsection{Measuring Convection}
Thermal convection is heat transport through fluid flow that is induced by temperature differences.
The strength of the local convective heat transport in the vertical direction is given by the product of the vertical velocity and the temperature:
\begin{equation} \label{eq:convective}
    q(x,y,t) = u_y(x,y,t) \theta(x,y,t) \,,
\end{equation}
where $\theta(x,y,t) = T(x,y,t) - \langle T \rangle_{x, y}$ is the temperature difference from the mean temperature over the field.
In the system, heat is transported by both convection and conduction. The ratio between transport by convection to conduction is an important measurement quantity known as the \textit{Nusselt number} Nu:
\begin{equation}\label{eq:nusselt}
    \Nu (t) = \frac{\langle q(x, y, t) \rangle_{x, y}}{\kappa (T_b - T_t) / H}\,.
\end{equation}
The numerator is a measure of the overall heat transport by convection (field average of \eqref{eq:convective}), whereas the denominator quantifies heat transport by conduction.\footnote{$H$: distance between top and lower boundary. $\kappa:$ thermal diffusivity.}
Note that with increasing $\Ra$, convective flows get stronger, which increases the value of Nu. \footnote{Nu scales like
$\text{Nu}_{\mathrm{Base}}(\Ra) \sim \Ra^{1/3}$
up to at least $Ra=10^{15}$ \cite{kartik_2020}.}

\subsection{Flow stabilization by controlling convection} \label{sec:methods:rbc_control}

\begin{figure*}
        \centering
        \begin{subfigure}[b]{0.59\textwidth}
        		\includegraphics[width=\textwidth]{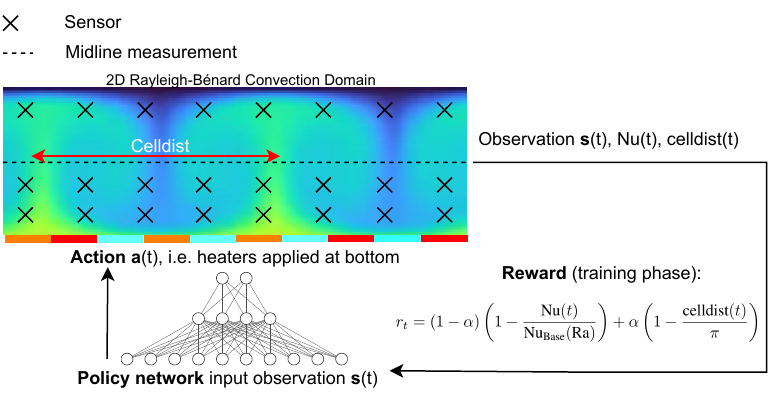}
        		\caption{}
        		\label{fig:control_setup}
        	\end{subfigure}
        	\hfill
        	\begin{subfigure}[b]{0.39\textwidth}
			\centering
		    \begin{subfigure}[b]{0.49\textwidth}
		        \centering
		        \includegraphics[width=\textwidth]{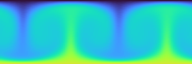}
		        \caption{}
		        \label{fig:example_ra10000}
		    \end{subfigure}
		    \begin{subfigure}[b]{0.49\textwidth}
		        \centering
		        \includegraphics[width=\textwidth]{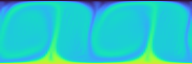}
		        \caption{}
		        \label{fig:example_ra1000000}
		    \end{subfigure}

		    \begin{subfigure}[b]{0.49\textwidth}
		        \centering
		        \includegraphics[width=\textwidth]{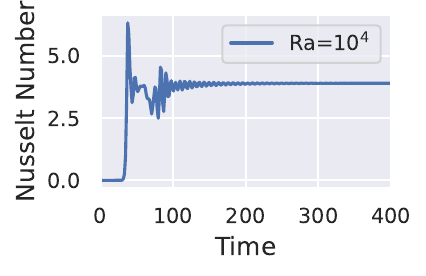}
		        \caption{}
		        \label{fig:baseline_episode_nusselt}
		    \end{subfigure}
		    \begin{subfigure}[b]{0.49\textwidth}
		        \centering
		        \includegraphics[width=\textwidth]{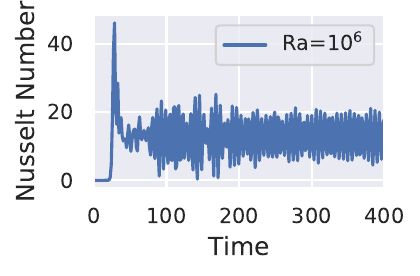}
		        \caption{}
		    \end{subfigure}        	
        	
        	\end{subfigure}
	\caption{Figure (a) shows the schematic of the control setup. The RL agent receives partial observations from an 8x48 grid of sensors and a midline vertical velocity measurement. Based on the observation $\bm{s}(t)$, the policy network outputs heating actions $\bm{a}(t)$ for 12 bottom actuators. Note in PPO, there is also a critic network involved that estimates the value function. The reward combines a normalized Nusselt number reduction term with a domain-informed term that encourages cell merging via the measured cell distance. Figures (b) and (d) show a state resulting from $Ra=10^4$ together with the Nusselt number over time. Figures (c) and (e) show the same for $Ra=10^6$.}
    \label{fig:controlsetup}
\end{figure*}
Motivated by interesting chaotic dynamics of the system and industrial relevance to control convective flows, the task is to robustly suppress convective heat transport, as measured by the Nusselt number in Eq.~\eqref{eq:nusselt}, by controlling heating elements at the bottom.
We illustrate the control setup in Fig.~\ref{fig:control_setup}, which bears some similarity to real-world lab set-ups such as \cite{howle1997}, where a simple PD control was applied.
In the control task, we divide the bottom boundary into 12 heating elements that each can be set to a temperature $a_i$.
Note that without further constraints, a trivial way to control the system is to set all heaters to the temperature at the top, i.e. $a_i = T_t$. This eliminates the temperature difference between bottom and top, so that heat transport (both conductive and convective) does not occur.
Hence, we consider a non-trivial scenario where the bottom heaters are constrained to values $a_i \in [T_b - 0.75, T_b + 0.75]$, which also simulates physical constraints of real-world heaters. In addition, we assume that the mean over all heaters is always $T_b$, which models scenarios in which the mean temperature difference between the top and bottom plate remains constant.\footnote{We satisfy the constraints by a simple transformation, see Supplementary Material.} Hence, the flow is to be controlled by temperature fluctuations applied by the heaters to the bottom boundary.

\subsubsection{Linear Control} \label{sec:methods:linear}
Since our objective is to study the practical feasibility of RL agents in chaotic settings, we consider a comparison to control schemes that are physically realizable and commonly used in practice, most notably proportional-derivative (PD) control, which is the standard approach in experimental setups \cite{howle1997,remillieux2007}.

The temperature fluctuations at the lower boundary are chosen to oppose the convective flow
by increasing heat below downward flows and decreasing heat below upward flows.
This is achieved by setting the error to a horizontal midline vertical velocity measurement, i.e. $E(x,t) = u_y(x,y=0,t)$ and computing the linear control signal:
\begin{flalign} \label{eq:pdcontrol}
a(x, t) = k_p E(x,t) + k_d  E'(x,t) \,,
\end{flalign}
where we found that gains of $k_p=-970$ and $k_d=-2000$ worked well across flow regimes.\footnote{To obtain the final $N=12$ heating values, we downsample the signal by averaging the grid points that correspond to a heater location and then apply a transformation to satisfy our control limits, see Supplementary Material.}

\subsubsection{Reinforcement Learning} \label{sec:methods:rl}
Although the mathematical formulation of the objective in RL bears similarity to that of conventional control, RL has the potential to discover sophisticated control policies through expressive neural representations, which is necessary for control of non-linear chaotic dynamics.
To train the agent, we employ the model-free algorithm PPO \cite{schulman2017}, a successful actor-critic method that improves robustness through a clipped objective. 
The agent learns a policy  $\pi(\bm{a}|\bm{s})$, which maps states $\bm{s}$ to action probabilities $\bm{a}$. The learning objective is to find a policy $\pi^*$ that maximizes the expected sum of reward,
$\max_{\pi} \mathbb{E} \left [ \sum_{t=0}^{\infty} \gamma^t R(\bm{s}_t,\bm{a}_t,\bm{s}_{t+1}) \right ]$,
where actions $\bm{a}_t$ are chosen according to $\pi$ and $0 \leq \gamma \leq 1$ is a discount factor that favors short-term over long-term rewards. The state transitions are modeled using a Markov Decision Process (MDP), which is in our case given by the underlying deterministic numerical simulation.

Fig.~\ref{fig:control_setup} gives an overview of the RL setup:
For the state observations, we assume access to probe sensors that are spread equidistantly on a $8 \times 48$ grid over the spatial domain and measure the local temperature and velocity of the fluid. We flatten all measurements into a vector with $3*8*48=1152$ elements as input to the actor and critic network.
The policy network outputs vectors $(a_1, a_2,\dots,a_{12}) \in \mathbb{R}^{12}$ (values for the 12 heaters), where $a_i \in [-1,1]$, which are transformed to satisfy a mean actuation of $T_b$ and limits of $[T_b - 0.75, T_b + 0.75]$, and then applied to the lower boundary.
As the agent's objective is to minimize convective heat transfer, which is measured by the Nusselt number $\Nu$ from Eq.~\eqref{eq:nusselt}, we incorporate $\Nu$ in the reward as follows:
\begin{flalign} \label{eq:reward_nusselt}
    R(s_t) = 1 - \frac{\text{Nu}(s_t)}{\text{Nu}_{\mathrm{Base}}(\Ra)} \, ,
\end{flalign}
where we additionally scaled $\Nu$ by $\text{Nu}_{\text{Base}}(\Ra)$, which we obtained as an average over the uncontrolled case, so that approximately $R(s_t) \in [0,1]$. PPO aims to maximize this reward over episodes, which corresponds to minimizing the Nusselt number.

\subsection{Reward Shaping: Better stabilization properties through cell merging} \label{sec:methods:rew_shaping}
Merging of B\'enard cells is an effective strategy for stabilizing flow and reducing convection, as was identified in \cite{vignon_2023} for the laminar flow at $\Ra=10^4$. Cell merging limits the number of counter-rotating flow structures in favor of a global, steady flow which moves slower. This results in lower Nusselt number with less variation over time.
However, cell merging is rather hard to achieve for chaotic flows, because of the existence of competing simple strategies that apply heating between cells. Although the simple strategies reduce the Nusselt number, the Nusselt number of single-cell states is lower and also associated with more steady flows.

In this work, we are interested in possibilities to facilitate RL for this challenging task and enhance the generalization ability of learned agents across initial conditions and flow scenarios by means of integration of prior knowledge. More specifically, we suggest to extend the reward function with domain-informed or physics-informed macroscopic quantities that can easily be measured:
In our case we focus on promoting B\'enard cell merging and study its effect on robust flow control. 
In addition to the $8 * 48$ grid of sensors, we assume a dense horizontal measurement of the vertical velocity field at the middle of the vertical axis, i.e. $u_y(x, y=0, t)$, inspired by the experiment in \cite{remillieux2007}.
We detect potential cell locations ${c_i}$ by finding positive peaks in this measurement.\footnote{We used find\_peaks from scipy.signal with height=0.}
In our setup, there are usually two convection cells with horizontal coordinate $c_1$ and $c_2$, and we can simply compute:
\begin{equation}
\text{celldist} = \min(|c_1 - c_2|, 2\pi - |c_1 - c_2|)\, ,
\end{equation}
which is the horizontal distance between the cells in the periodic domain $x \in [0, 2\pi]$.
In the general case of an arbitrary number of cells, we simply compute the maximum of the pair-wise cell distances to summarize the overall degree of cell merging.
Next, we modify the reward function to include the cell distance as follows:
\begin{equation} \label{eq:reward_shaping_reward}
    r_t = (1-\alpha) \left( 1 - \frac{\Nu(t)}{\Nu_{\text{Base}}(\Ra)} \right)  + \alpha \left (1 - \frac{\text{celldist}(t)}{\pi} \right) \,,
\end{equation}
where $\alpha \in [0,1]$ balances the cell distance and the Nusselt number in the reward. The quantity $(1-\text{celldist}(t)/\pi)$ ranges from $0$, when the cells are maximally separated,\footnote{Note that $\pi$ is the maximum possible distance on the periodic domain $x \in [0, 2\pi]$.} to $1$, in case of a single merged cell.

\section{Experiments and Results}
We evaluate the effect of the domain-informed RL agents on generalization performance in three experiments: In Experiment 1, we exclude domain knowledge by using $\alpha=0$ in Eq.~\eqref{eq:reward_shaping_reward} and term the resulting agents as \textit{uninformed}. In Experiment 2, we study the effect of including the domain knowledge, i.e. $\alpha > 0$, and we call the resulting agents \textit{domain-informed}. In Experiment 3, we explicitly study for both cases the generalization performance to other flow regimes by using the agents that were trained on $Ra=10^5$ to control the flow at $Ra=10^4$ and $Ra=10^6$.
In all experiments, we evaluate the agent at an increasing $\Ra$ (i.e. increasing chaotic convection) across a variety of initial conditions.

\subsection{Episode Rollouts and PPO training}
We trained agents on Rayleigh numbers $\Ra \in \{10^4,10^5,10^6,5\cdot10^6\}$.
Each simulation starts from a no-motion state with small random perturbations in the vertical temperature gradient. Due to equivariance of the dynamics under horizontal shifts, these small random perturbations lead to random horizontal shifts in the obtained cell configurations. For each Ra, we generated 37 initial conditions in this manner.

During training, the agent is exposed to 20 of these convective states. During training, we validate the agent on 5 unseen states and save the best performing agent. Final evaluation is performed on the remaining 12 independent convective states not seen during training or validation. This setup provides a strong test of the agent's generalization capabilities.

We use the PPO implementation from \cite{sb3} and collect training data from multiple parallel rollouts. A full description of PPO training parameters (number of rollouts, training times, etc) is provided in the SM.

\begin{figure}
    \centering
    \begin{subfigure}[t]{0.49\linewidth}
        \centering
        \includegraphics[width=\textwidth]{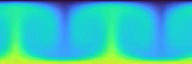}
        \caption{$t=0$}
        \label{fig:exp1:t0}
    \end{subfigure}
    \hfill
    \begin{subfigure}[t]{0.49\linewidth}
        \centering
        \includegraphics[width=\textwidth]{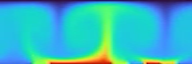}
        \caption{$t=12$}
        \label{fig:exp1:t12}
    \end{subfigure}
    \begin{subfigure}[t]{0.49\linewidth}
        \centering
        \includegraphics[width=\textwidth]{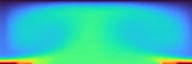}
        \caption{$t=60$}
        \label{fig:exp1:t60}
    \end{subfigure}
    \hfill
    \begin{subfigure}[t]{0.49\linewidth}
        \centering
        \includegraphics[width=\textwidth]{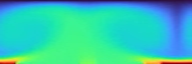}
        \caption{$t=210$}
        \label{fig:exp1:t210}
    \end{subfigure}
    \begin{subfigure}[t]{0.49\linewidth}
        \centering
        \includegraphics[width=\textwidth]{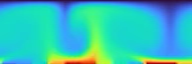}
        \caption{$t=270$}
        \label{fig:exp1:t270}
    \end{subfigure}
    \hfill
    \begin{subfigure}[t]{0.49\linewidth}
        \centering
        \vspace{-1.5cm}
        \includegraphics[width=\textwidth]{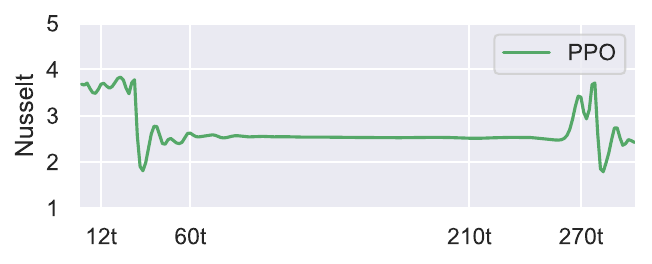}
        
        \caption{Nusselt Reduction of rollout}
        \label{fig:exp1:ppo_1e4:nu}
    \end{subfigure}
    \caption{Flow control by an \textit{uninformed agent} for a typical test set episode at $\Ra = 10^4$.}
    \label{fig:exp1:ppo_1e4}
\end{figure}

\subsection{Experiment 1: Uninformed RL-based flow control} \label{sec:experiment1}
\begin{figure}
	\centering
	\includegraphics[width=\columnwidth]{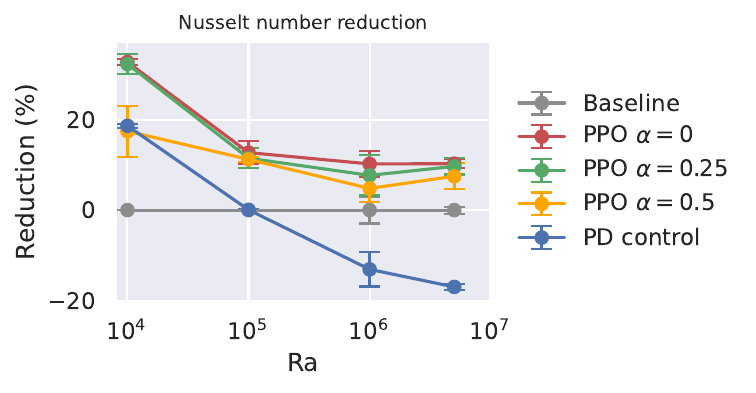}
	\caption{The relative reduction of the Nusselt number with respect to the uncontrolled baseline for each control method on each Ra. For each test checkpoint we computed the mean Nusselt number over the entire episode and then compute the percentage change relative to the mean Nusselt number of the uncontrolled baseline. Here, we show the average and standard deviation computed over the 12 test checkpoints.}
	\label{fig:exp1_exp2_exp3_nusselt_reduction}
\end{figure}
We first set a baseline by evaluating the performance of uninformed RL agents in reducing convective flow using $\alpha=0$ in Eq.~\eqref{eq:reward_shaping_reward}. 
In Fig.~\ref{fig:exp1_exp2_exp3_nusselt_reduction}, three results stand out: (1) The uninformed RL agent ($\alpha=0$) shows significant promise in reducing the convective flow, as for all considered Ra it reduces the Nusselt number significantly. (2) The standard deviation is very low, meaning a consistent Nusselt number reduction is achieved across all test checkpoints. (3) PD control, which is used in lab experiments, can only achieve satisfactory performance for the laminar flow at $Ra=10^4$.

For the laminar flow at $Ra=10^4$, we show in Fig.~\ref{fig:exp1:ppo_1e4} the temperature field over time when controlled by the uninformed agent.
In Fig.~\ref{fig:exp1:t0}, we see that the uninformed agent starts to merge the left cell with the right cell.
After the merge, the agent gradually widens the single cell (Fig.~\ref{fig:exp1:t60} and Fig.~\ref{fig:exp1:t210}), which it identified as a strategy to reduce the overall convective flow.
This widening results in a split of the single-cell into the two-cell configuration again (Fig.~\ref{fig:exp1:t270}). Hence, the flow is not stabilized under the control.
In Fig.~\ref{fig:exp1:ppo_1e4:nu}, we marked the time points of the snapshots on the x-axis: The cell merge is associated with a significant decrease in Nu which then becomes stationary. Later in the episode, the brief split into two cells at $t=270$ is associated with an instability in the Nusselt number.

For $Ra>10^4$, successful control was almost never achieved, because the uninformed agent remained stuck in simpler strategies that amount to heating between the cells.
Although we observed that more exploration can somewhat alleviate the issue, this requires much more training effort.
These baseline results highlight that without domain knowledge, stabilization of chaotic flows is not consistently achieved, motivating the need for domain-informed priors.

\subsection{Experiment 2: Exploring stabilization properties by adding domain knowledge} \label{sec:experiment3}

We experiment with the inclusion of domain knowledge through promoting B\'enard cell merging using the reward function in Eq.~\eqref{eq:reward_shaping_reward} ($\alpha>0$).
For the coefficient $\alpha$ that determines importance of this term, we initially experimented with values $\alpha=0.25$ and $\alpha=0.5$.

As shown in Fig.~\ref{fig:exp1_exp2_exp3_nusselt_reduction}, the value $\alpha=0.25$ resulted in robust performance in Nusselt Number reduction across all Ra. This is promising, as a sensitive dependence on $\alpha$ would hamper practical deployments. We use $\alpha=0.25$ in the rest of the discussion.

\begin{figure}
     \centering
     \begin{subfigure}[b]{0.49\columnwidth}
         \centering
         \includegraphics[width=\textwidth]{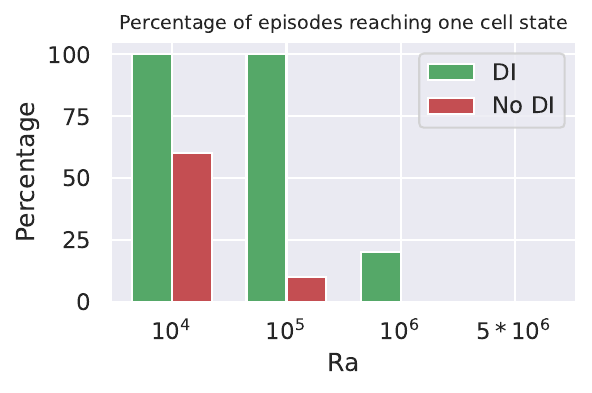}
         \caption{}
         \label{fig:exp3:percentage_episodes_one_cell}
     \end{subfigure}
     \hfill
     \begin{subfigure}[b]{0.49\columnwidth}
         \centering
         \includegraphics[width=\textwidth]{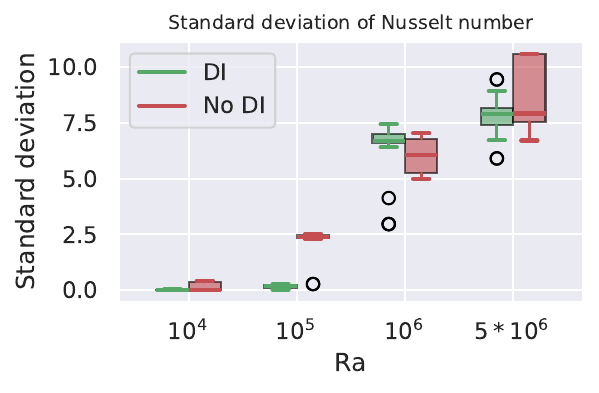}
         \caption{}
         \label{fig:exp3:reduction_nusselt_std}
     \end{subfigure}
     \begin{subfigure}[b]{0.49\columnwidth}
         \centering
         \includegraphics[width=\textwidth]{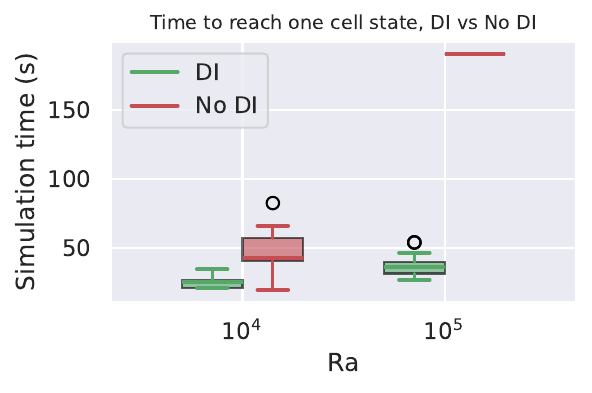}
         \caption{}
         \label{fig:exp3:time_eps_onecell}
     \end{subfigure}
     \hfill
     \begin{subfigure}[b]{0.49\columnwidth}
         \centering
         \includegraphics[width=\textwidth]{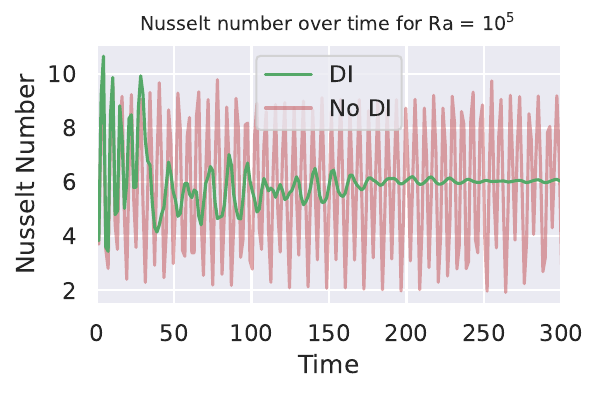}
         \caption{}
         \label{fig:exp3:nusselt_over_time}
     \end{subfigure}
     \caption{The effect of Domain-Informed (DI) training vs. uninformed training (No DI) on flow control shown by three key statistics computed on the test set. (a): the percentage of episodes where B\'enard cells were merged. (b): The standard deviation of the Nusselt number over time, computed over the last 40 actions in the episode. (c): The simulation time at which the cell merging event took place. (d) The Nusselt number during a typical example test episode.}
     \label{fig:exp3:stats}
\end{figure}

The key advantage of domain-informed RL is its ability to consistently stabilize initially chaotic flows, such as at $Ra = 10^5$, into a steady state.
We summarize the main differences using key statistics in Fig.~\ref{fig:exp3:stats}: Domain-informed training always merged cells in the regime $Ra=10^4$ and $Ra=10^5$, as shown in Fig.~\ref{fig:exp3:percentage_episodes_one_cell}. In addition, this resulted in consistent flow stabilization in this regime (Fig.~\ref{fig:exp3:reduction_nusselt_std}).
This is significant, as for $Ra=10^5$ the uncontrolled flow is chaotic and became consistently stable under the domain-informed control\footnote{The videos (link under abstract) clearly illustrate this behavior}.
To illustrate the flow stabilization further, in Fig.~\ref{fig:exp3:nusselt_over_time}, examples of the Nusselt number during typical episodes of the Domain-Informed agent (in green) and the uninformed agent (in red), where the domain-informed agent achieves a stationary Nusselt number.
Figures \ref{fig:exp3:time_eps_onecell} and Fig.~\ref{fig:exp3:reduction_nusselt_std} further summarize the obtained flows: In \ref{fig:exp3:time_eps_onecell} we show that the domain-informed agents achieve cell merging earlier in the episodes\footnote{As cell merging was not consistently achieved for the uninformed agent, we only include the episodes here in which it was achieved.}. Fig.~\ref{fig:exp3:reduction_nusselt_std} confirms the result in Fig.~\ref{fig:exp3:nusselt_over_time} that stationary flows were consistently achieved for $Ra=10^4$ and $Ra=10^5$. For the regime $Ra>10^5$, the differences between the domain-informed and the uninformed control are minor: in this highly chaotic regime, both agents resort to the simple control strategy that still reduces the Nusselt number (see Fig.~\ref{fig:exp1_exp2_exp3_nusselt_reduction}).
In summary, in the regime $Ra=10^5$, the domain-informed agent consistently transformed a chaotic flow into a stable single-cell configuration with constant Nusselt number, a capability absent in uninformed agents.

In the SM, we give training curves of domain-informed and uninformed agents for $Ra=10^4, Ra=10^5$.
They show that the domain-informed agent achieves flow stabilization early in training. We also show that the uninformed agent tends to overfit to the training set. In contrast, the domain-informed training objective is less prone to over-fitting and achieves much better generalization to unseen conditions beyond the training set.

\section{Experiment 3: Generalization to different flow regimes}
\begin{figure}
     \centering
     \begin{subfigure}[b]{0.49\columnwidth}
         \centering
         \includegraphics[width=\textwidth]{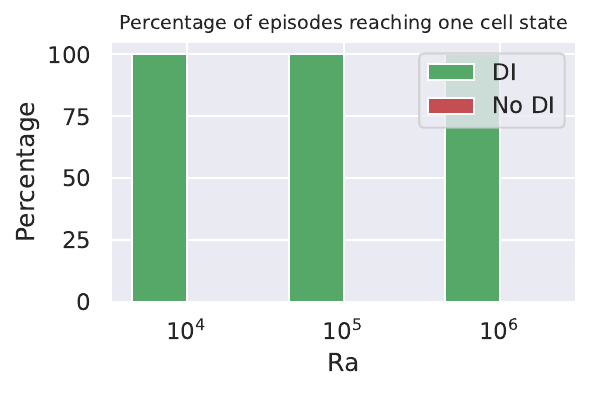}
         \caption{}
         \label{fig:exp4:percentage_episodes_one_cell_gen}
     \end{subfigure}
     \hfill
     \begin{subfigure}[b]{0.49\columnwidth}
         \centering
         \includegraphics[width=\textwidth]{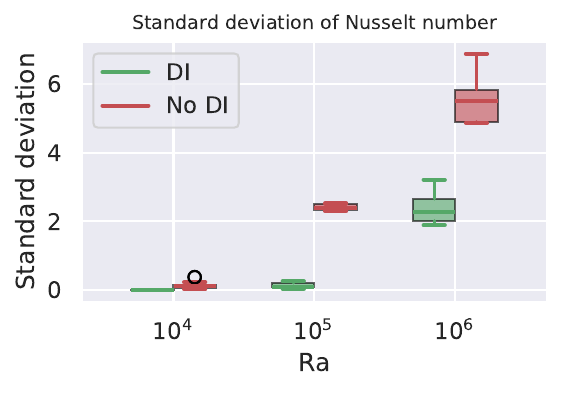}
         \caption{}
         \label{fig:exp4:reduction_nusselt_std}
     \end{subfigure}
     \begin{subfigure}[b]{0.49\columnwidth}
         \centering
         \includegraphics[width=\textwidth]{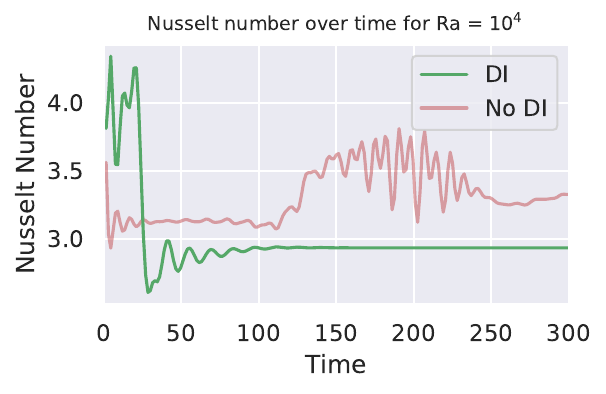}
         \caption{}
         \label{fig:exp4:nusselt_over_time_RS_NoRS_Ra_104_gen}
     \end{subfigure}
     \hfill
     \begin{subfigure}[b]{0.49\columnwidth}
         \centering
         \includegraphics[width=\textwidth]{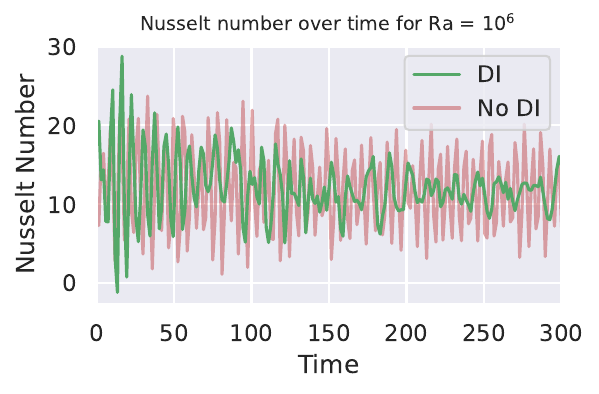}
         \caption{}
         \label{fig:exp4:nusselt_over_time_RS_NoRS_Ra_106_gen}
     \end{subfigure}
     \caption{The effect of domain-informed training of an agent trained on $Ra=10^5$ on generalization across other $Ra$ as well, in three key statistics computed on the test set. See Fig.~\ref{fig:exp3:stats} for an explanation of the statistics shown in (a) and (b). Fig.~(c) and (d): The Nusselt number during a typical example test episode for $Ra=10^4$ and $Ra=10^6$, respectively.}
     \label{fig:exp4:stats_gen}
\end{figure}

We used the domain-informed agent trained on flows at $Ra=10^5$ for controlling flows at $Ra=10^4$ and $Ra=10^6$ and we compared with the same scenario using the uninformed agent. Fig.~\ref{fig:exp4:stats_gen} shows that the domain-informed agent has consistent success in achieving cell merging and associated flow stability, whereas the uninformed agent does not achieve these properties.
Counter-intuitively, the domain-informed performance on $Ra=10^6$ is better than when training on $Ra=10^6$ (cf. Fig.~\ref{fig:exp3:stats}). We attribute this to the fact that training of policies in lower chaotic regimes is easier, and such policies may therefore show better generalization ability to  more highly chaotic regimes. The success of the domain-informed agents in cell merging resulted in stationary Nusselt numbers for $Ra=10^4$ (Fig.~\ref{fig:exp4:nusselt_over_time_RS_NoRS_Ra_104_gen} green line), and a significant reduction for $Ra=10^6$ (Fig.~\ref{fig:exp4:nusselt_over_time_RS_NoRS_Ra_106_gen} green line), underlining that the learned stabilization strategy generalizes to other flow regimes including higher levels of chaos.

\section{Discussion}

Our results provide clear evidence that the inclusion of domain knowledge is a critical component to achieve robust control and stabilization of chaotic fluid flow. A key insight is that elegant domain-informed reward design plays a critical role in enabling robust flow control, obtaining steady flows in regimes where the uncontrolled flow is chaotic.
Although there are slight differences in the sensor setup between the domain-informed and the uninformed cases, the cell distance can also be well-approximated from the dense sensors in the horizontal direction used for the uninformed case, which would not change the results.
Moreover, our results indicated that a balancing value of $\alpha=0.25$ in the reward in Eq.~\eqref{eq:reward_shaping_reward} already resulted in robust performance across a large flow regime, making fine-tuning not critical, which is a major advantage for physical implementation.

For the uninformed agent, we only obtained steady flows for $Ra=10^4$, but this state was not stable (Fig.~\ref{fig:exp1:t270}) and achieved late in training (See SM).
Encouraging B\'enard cell merging was an effective way to guide the agent away from the simple two-cell control strategy towards a one-cell setup that has lower convective heat transfer. In addition, those one-cell setups were associated with steady flows: A key result is that for $Ra=10^5$, which is a regime with chaotic flow without control, the domain-informed agent consistently merged cells and transformed the chaotic flow into a stable, steady flow with constant Nusselt number (e.g., see Fig. 4d). The domain-informed agent also exhibited generalization ability to flows up to $Ra=10^6$.

\subsection{Conclusion}
This work bridges Artificial Intelligence and Engineering by demonstrating how domain-informed RL can enable robust control of chaotic flows.
Through elegant domain-informed reward design, our agents learned significantly more robust control across initial conditions and exhibited a level of generalization ability across flow regimes.
These insights highlight the potential of domain-informed RL for effective robust control in complex flow regimes, especially in settings where conventional control fails.

A challenging next step is the control of 3D convective flows, including turbulent effects. As our experiments in 2D indicated that including domain knowledge becomes increasingly important for complex flows where control is feasible, the inclusion of domain-informed rewards would be a critical factor for enabling flow control in 3D. Our proposed reward is strongly tied to 2D, but similar domain-informed counterparts could be formulated in 3D. We also aim to further reduce sample requirements by learning surrogate models of the dynamics, which allows for agent training using the surrogate model instead of costly rollouts of the environment. A starting point would be to extend our uncontrolled surrogates for RBC in \cite{markmann2024,straat_2025} with control. Eventually, it would be interesting to explore the use of robust domain-informed agents for the control of real-world convective flows in lab experiments.

\section{Acknowledgements}
The authors acknowledge financial support by the project ”SAIL: SustAInable Life-cycle of Intelligent Socio-Technical Systems” (Grant ID NW21-059A and NW21-059D), which is funded by the program ”Netzwerke 2021” of the Ministry of Culture and Science of the State of North Rhine Westphalia, Germany. SP acknowledges support by the European Union via the ERC Starting Grant ``KoOpeRaDE'' (Grant ID 101161457).



\appendix

\bibliography{aaai2026}

\begin{thebibliography}{19}
\providecommand{\natexlab}[1]{#1}

\bibitem[{Banerjee et~al.(2025)Banerjee, Nguyen, Fookes, and Raissi}]{banerjee_2025}
Banerjee, C.; Nguyen, K.; Fookes, C.; and Raissi, M. 2025.
\newblock A survey on physics informed reinforcement learning: {Review} and open problems.
\newblock \emph{Expert Systems with Applications}, 287: 128166.

\bibitem[{Beintema et~al.(2020)Beintema, Corbetta, Biferale, and Toschi}]{beintema_2020}
Beintema, G.; Corbetta, A.; Biferale, L.; and Toschi, F. 2020.
\newblock Controlling Rayleigh–B\'enard convection via reinforcement learning.
\newblock \emph{Journal of Turbulence}, 21(9-10): 585--605.

\bibitem[{Garnier et~al.(2021)Garnier, Viquerat, Rabault, Larcher, Kuhnle, and Hachem}]{garnier_2021}
Garnier, P.; Viquerat, J.; Rabault, J.; Larcher, A.; Kuhnle, A.; and Hachem, E. 2021.
\newblock A review on deep reinforcement learning for fluid mechanics.
\newblock \emph{Computers \& Fluids}, 225: 104973.

\bibitem[{Gu, Hegde, and Bishop(2018)}]{gu2018}
Gu, Y.; Hegde, V.; and Bishop, K. J.~M. 2018.
\newblock Measurement and mitigation of free convection in microfluidic gradient generators.
\newblock \emph{Lab Chip}, 18(22): 3371--3378.
\newblock Publisher: The Royal Society of Chemistry.

\bibitem[{Howle(1997)}]{howle1997}
Howle, L.~E. 1997.
\newblock Active Control of {{Rayleigh}}--{{B{\'e}nard}} Convection.
\newblock \emph{Physics of Fluids}, 9: 1861–1863.

\bibitem[{Hsia and Nishida(2022)}]{hsia_route_2022}
Hsia, C.-H.; and Nishida, T. 2022.
\newblock A {Route} to {Chaos} in {Rayleigh}–{B\'enard} {Heat} {Convection}.
\newblock \emph{Journal of Mathematical Fluid Mechanics}, 24(2): 38.

\bibitem[{Iyer et~al.(2020)Iyer, Scheel, Schumacher, and Sreenivasan}]{kartik_2020}
Iyer, K.~P.; Scheel, J.~D.; Schumacher, J.; and Sreenivasan, K.~R. 2020.
\newblock Classical 1/3 scaling of convection holds up to $Ra = 10^{15}$.
\newblock \emph{Proceedings of the National Academy of Sciences}, 117(14): 7594--7598.

\bibitem[{Markmann, Straat, and Hammer(2024)}]{markmann2024}
Markmann, T.; Straat, M.; and Hammer, B. 2024.
\newblock Koopman-Based Surrogate Modelling of Turbulent Rayleigh-B\'enard Convection.
\newblock In \emph{2024 International Joint Conference on Neural Networks (IJCNN)}, 1--8.

\bibitem[{Mortensen(2018)}]{mortensen_joss}
Mortensen, M. 2018.
\newblock Shenfun: High performance spectral Galerkin computing platform.
\newblock \emph{Journal of Open Source Software}, 3(31): 1071.

\bibitem[{Pandey, Scheel, and Schumacher(2018)}]{pandey2018}
Pandey, A.; Scheel, J.~D.; and Schumacher, J. 2018.
\newblock Turbulent superstructures in {Rayleigh}-{B\'enard} convection.
\newblock \emph{Nature Communications}, 9(2118).

\bibitem[{Raffin et~al.(2021)Raffin, Hill, Gleave, Kanervisto, Ernestus, and Dormann}]{sb3}
Raffin, A.; Hill, A.; Gleave, A.; Kanervisto, A.; Ernestus, M.; and Dormann, N. 2021.
\newblock Stable-Baselines3: Reliable Reinforcement Learning Implementations.
\newblock \emph{Journal of Machine Learning Research}, 22(268): 1--8.

\bibitem[{Remillieux, Zhao, and Bau(2007)}]{remillieux2007}
Remillieux, M.~C.; Zhao, H.; and Bau, H.~H. 2007.
\newblock Suppression of Rayleigh-B\'enard convection with proportional-derivative controller.
\newblock \emph{Physics of Fluids}, 19(1): 017102.

\bibitem[{Schulman et~al.(2017)Schulman, Wolski, Dhariwal, Radford, and Klimov}]{schulman2017}
Schulman, J.; Wolski, F.; Dhariwal, P.; Radford, A.; and Klimov, O. 2017.
\newblock Proximal {{Policy Optimization Algorithms}}.
\newblock arXiv:1707.06347.

\bibitem[{Straat, Markmann, and Hammer(2025)}]{straat_2025}
Straat, M.; Markmann, T.; and Hammer, B. 2025.
\newblock Solving Turbulent Rayleigh-B\'enard Convection using Fourier Neural Operators.
\newblock In \emph{Proceedings of the 33rd European Symposium on Artificial Neural Networks, Computational Intelligence and Machine Learning (ESANN 2025)}. Bruges, Belgium: i6doc.com.
\newblock ISBN 9782875870933.

\bibitem[{Tang and Bau(1993)}]{tang_1993}
Tang, J.; and Bau, H.~H. 1993.
\newblock Stabilization of the no-motion state in Rayleigh-B\'enard convection through the use of feedback control.
\newblock \emph{Phys. Rev. Lett.}, 70: 1795--1798.

\bibitem[{Vasanth et~al.(2024)Vasanth, Rabault, {Alc{\'a}ntara-{\'A}vila}, Mortensen, and Vinuesa}]{vasanth2024}
Vasanth, J.; Rabault, J.; {Alc{\'a}ntara-{\'A}vila}, F.; Mortensen, M.; and Vinuesa, R. 2024.
\newblock Multi-Agent {{Reinforcement Learning}} for the {{Control}} of {{Three-Dimensional Rayleigh}}--{{B{\'e}nard Convection}}.
\newblock \emph{Flow, Turbulence and Combustion}, 1--37.

\bibitem[{Vignon et~al.(2023)Vignon, Rabault, Vasanth, Alcántara-Ávila, Mortensen, and Vinuesa}]{vignon_2023}
Vignon, C.; Rabault, J.; Vasanth, J.; Alcántara-Ávila, F.; Mortensen, M.; and Vinuesa, R. 2023.
\newblock Effective control of two-dimensional Rayleigh–B\'enard convection: Invariant multi-agent reinforcement learning is all you need.
\newblock \emph{Physics of Fluids}, 35(6): 065146.

\bibitem[{Vinuesa, Brunton, and McKeon(2023)}]{vinuesa_transformative_2023}
Vinuesa, R.; Brunton, S.~L.; and McKeon, B.~J. 2023.
\newblock The transformative potential of machine learning for experiments in fluid mechanics.
\newblock \emph{Nature Reviews Physics}, 5(9): 536--545.

\bibitem[{Wang et~al.(2024)Wang, Cao, Huang, Liu, Hu, Luo, Song, Zhao, Liu, Sun, Zhang, Wei, Wang, Wu, Ma, and Sun}]{wang_2024}
Wang, H.; Cao, Y.; Huang, Z.; Liu, Y.; Hu, P.; Luo, X.; Song, Z.; Zhao, W.; Liu, J.; Sun, J.; Zhang, S.; Wei, L.; Wang, Y.; Wu, T.; Ma, Z.-M.; and Sun, Y. 2024.
\newblock Recent {Advances} on {Machine} {Learning} for {Computational} {Fluid} {Dynamics}: {A} {Survey}.
\newblock ArXiv:2408.12171 [cs].

\end{thebibliography}

\end{document}